\documentclass[10pt,twocolumn,letterpaper]{article}

\usepackage{iccv}
\usepackage{times}
\usepackage{epsfig}

\usepackage{graphicx}
\graphicspath{{figure/}}
\usepackage{pythonhighlight}
\usepackage{amsmath}
\usepackage{amssymb}
\usepackage{subfigure}
\usepackage{listings}
\usepackage{multirow}

\usepackage{makecell}
\usepackage{fdsymbol}
\usepackage[bold]{hhtensor}
\usepackage{algpseudocode}
\usepackage{algorithmicx,algorithm}

\usepackage{booktabs}       
\usepackage{microtype}      
\usepackage[american]{babel}

\usepackage{pifont}

\usepackage[pagebackref=true,breaklinks=true,colorlinks,bookmarks=false]{hyperref}

\newcommand{\squishlist}{
 \begin{list}{$\bullet$}
  { \setlength{\itemsep}{0pt}
     \setlength{\parsep}{1pt}
     \setlength{\topsep}{1pt}
     \setlength{\partopsep}{0pt}
     \setlength{\leftmargin}{1em}
     \setlength{\labelwidth}{1em}
     \setlength{\labelsep}{0.5em} } }

\newcommand{\squishend}{
  \end{list}  }

\usepackage{pifont}

\usepackage[breaklinks=true,bookmarks=false]{hyperref}

\iccvfinalcopy

\def\iccvPaperID{****}

\ificcvfinal\fi

\begin{document}

\title{Residual Attention: A Simple but Effective Method for Multi-Label Recognition}

\author{Ke Zhu \quad Jianxin Wu\thanks{This research was partly supported by the National Natural Science Foundation of China under Grant 61772256 and 61921006.}\\
State Key Laboratory for Novel Software Technology\\
Nanjing University, China \\
{\tt\small zhuk@lamda.nju.edu.cn, wujx2001@nju.edu.cn}
}

\maketitle
\ificcvfinal\thispagestyle{empty}\fi

\begin{abstract}
	Multi-label image recognition is a challenging computer vision task of practical use. Progresses in this area, however, are often characterized by complicated methods, heavy computations, and lack of intuitive explanations. To effectively capture different spatial regions occupied by objects from different categories, we propose an embarrassingly simple module, named class-specific residual attention (CSRA). CSRA generates class-specific features for every category by proposing a simple spatial attention score, and then combines it with the class-agnostic average pooling feature. CSRA achieves state-of-the-art results on multi-label recognition, and at the same time is much simpler than them. Furthermore, with only 4 lines of code, CSRA also leads to consistent improvement across many diverse pretrained models and datasets without any extra training. CSRA is both easy to implement and light in computations, which also enjoys intuitive explanations and visualizations.
\end{abstract}

\section{Introduction}

Convolutional neural networks (CNNs) have dominated many computer vision tasks, especially in image classification. However, although many network architectures have been proposed for single-label classification, e.g., VGG~\cite{VGG}, ResNet~\cite{ResNet}, EfficientNet~\cite{Efficientnet} and VIT~\cite{VIT}, the progress in multi-label recognition remains modest. In multi-label tasks, the objects' locations and sizes vary a lot and it is difficult to learn a single deep representation that fits all of them.

Recent studies in multi-label recognition mainly focus on three aspects: semantic relations among labels, object proposals, and attention mechanisms. To explore semantic relations, Bayesian network~\cite{2011_IJCAI_Markov,2016_CVPR_condition_graph}, Recurrent Neural Network (RNN)~\cite{2017_CVPR_RNN,2016_CVPR_RNNCNN} and Graph Convolutional Network (GCN)~\cite{2019_CVPR_GCN,2019_ICCV_GCNre} have been adopted, but they suffer from high computational cost or manually defined adjacency matrices. Proposal based methods~\cite{2015_PAMI_HCP,2016_ICIP_RCP,Wider} spend too much time in processing the object proposals. Although attention models are end-to-end and relatively simple, for multi-label classification they resort to complicated spatial attention models~\cite{2018_ECCV_CAM,2017_CVPR_SRN,2020_arxiv_Gaobb}, which are difficult to optimize, implement, or interpret. Instead, we propose an embarrassingly simple and easy to train \emph{class-specific residual attention} (CSRA) module to fully utilize the spatial attention for each object class separately, and achieves superior accuracy. The CSRA module has negligible computational cost, too.

Our motivation came from Fig.~\ref{fig:pytorch code}, in which \emph{only 4 lines of code consistently leads to improvement of multi-label recognition, across many diverse pretrained models and datasets, even \emph{without} any extra training}, as detailed in Table~\ref{tab:free improvement}. The only change is to add a global max-pooling on top of the usual global average pooling, but the improvement is consistent. Its advantage is also verified on ImageNet, a single-label recognition task.

\begin{figure}
\begin{python}
  # x: feature tensor, output of CNN backbone
  # x's size: (B, d, H, W)
  # y_raw: by applying classifier ('FC') to 'x'
  # y_raw's size: (B, C, HxW)
  # C: number of classes
  y_raw = FC(x).flatten(2)
  y_avg = torch.mean(y_raw, dim=2)
  y_max = torch.max(y_raw, dim=2)[0]
  score = y_avg + Lambda * y_max
\end{python} 
\caption{A simple modification in the testing stage using PyTorch, in which \pyth{Lambda} (or $\lambda$) is a hyperparameter that combines global average and max pooling scores. When \pyth{Lambda} ($\lambda$) is 0, \pyth{score} equals \pyth{y_avg}, which is the score of the baseline model.}
\label{fig:pytorch code}
\end{figure}

\begin{table*}
	\caption{An embarrassingly simple, almost zero-cost, and training-free improvement to a diverse set of existing models on 3 multi-label recognition datasets and 1 single-label recognition dataset.}
	\label{tab:free improvement}
	\centering
	\small
	\begin{tabular}{c|cc|ccc|ccc} 
		\hline
		\thead{Datasets \&\\resolutions} & Models & \thead{Baseline\\mAP/acc} & \thead{Varying\\$\lambda$} & \thead{mAP /\\acc} & \thead{mAP / acc\\difference} & \thead{Fixed\\$\lambda$} & \thead{mAP /\\acc} & \thead{mAP / acc\\difference}\\
		\hline \hline
		\multirow{6}{*}{\thead{MS-COCO\\448$\times$448}}  & MobileNet~\cite{MobileNet}    & 71.8 & 0.2       & \textbf{73.0}  &  1.2 \textuparrow & 0.2  & \textbf{73.0}  & 1.2 \textuparrow  \\
													   & EfficientNet~\cite{Efficientnet} & 75.6 & 0.1       & \textbf{76.1}  & 0.5 \textuparrow & 0.2  & \textbf{76.1}  & 0.5 \textuparrow  \\
													   & VIT-B16-224~\cite{VIT}     & 79.0 & 0.8       & \textbf{79.7}  & 0.7 \textuparrow & 0.2  & \textbf{79.3}  & 0.3 \textuparrow    \\
													   & VIT-L16-224~\cite{VIT}     & 80.4 & 0.4       & \textbf{80.6}  & 0.2 \textuparrow & 0.2  & \textbf{80.6}  & 0.2 \textuparrow  \\
													   & ResNet-cut~\cite{ResNet}         & 82.4 & 0.05      & \textbf{82.6}  & 0.2 \textuparrow & 0.02 & \textbf{82.5}  & 0.1 \textuparrow  \\
													   & GCN~\cite{2019_CVPR_GCN}         & 83.0 & 0.2       & \textbf{83.2}  & 0.2 \textuparrow & 0.2  & \textbf{83.2}  & 0.2 \textuparrow \\
		\hline
		\multirow{3}{*}{\thead{VOC2007\\448$\times$448}}   & MobileNet~\cite{MobileNet}       & 89.6 & 0.2       & \textbf{90.3}  & 0.7 \textuparrow & 0.2  & \textbf{90.3}  & 0.7 \textuparrow \\
													   & ResNet-cut~\cite{ResNet}         & 93.9 & 0.05      & \textbf{94.0}  & 0.1 \textuparrow & 0.02 & \textbf{94.0}  & 0.1 \textuparrow \\
													   & GCN~\cite{2019_CVPR_GCN}         & 94.0 & 0.2       & \textbf{94.1}  & 0.1 \textuparrow & 0.2  & \textbf{94.1}  & 0.1 \textuparrow \\
		\hline
		\multirow{3}{*}{\thead{WIDER\\224$\times$224}} & MobileNet~\cite{MobileNet}       & 72.1 & 0.1       & \textbf{72.3}  & 0.2 \textuparrow & 0.2  & \textbf{72.2}  & 0.1 \textuparrow \\
													   & VIT-B16-224~\cite{VIT}     & 86.3 & 0.3       & \textbf{86.4}  & 0.1 \textuparrow & 0.2  & \textbf{86.4}  & 0.1 \textuparrow \\
													   & VIT-L16-224~\cite{VIT}     & 87.7 & 0.2       & \textbf{87.8}  & 0.1 \textuparrow & 0.2  & \textbf{87.8}  & 0.1 \textuparrow \\
			\hline
		\multirow{4}{*}{ImageNet}                      & ResNet-50~\cite{ResNet}          & 75.6 & 0.03      & \textbf{75.7}  & 0.1 \textuparrow & 0.02 & \textbf{75.7}  & 0.1 \textuparrow \\
													   & ResNet-101~\cite{ResNet}         & 77.1 & 0.02      & \textbf{77.2}  & 0.1 \textuparrow & 0.02 & \textbf{77.2}  & 0.1 \textuparrow \\
													   & VIT-B16-224~\cite{VIT}           & 80.5 & 0.2       & \textbf{80.7}  & 0.2 \textuparrow & 0.2  & \textbf{80.7}  & 0.2 \textuparrow \\
													   & VIT-B16-384~\cite{VIT}           & 83.5 & 0.3       & \textbf{83.6}  & 0.1 \textuparrow & 0.2  & \textbf{83.6}  & 0.1 \textuparrow \\    
		\hline
	\end{tabular}
 \end{table*}

In this paper, we show that this operation, the max pooling among different spatial regions for every class, is in fact a class-specific attention operation, which can be further viewed as a residual component of the class-agnostic global average pooling. Hence, we generalize it to propose a simple class-specific residual attention module (CSRA), and has achieved state-of-the-art performance on four multi-label datasets, namely VOC2007~\cite{VOC}, VOC2012~\cite{VOC2012}, MS-COCO~\cite{MSCOCO} and WIDER-Attribute~\cite{Wider}. Furthermore, the proposed CSRA has an intuitive explanation on how spatial attention is integrated into it.

Our contributions can be summarized as:
\begin{itemize}
	\item An extremely simple but effective method to improve pretrained models without any further training;
	\item A simple and effective CSRA module that achieves excellent results on four multi-label recognition datasets;
	\item An intuitive interpretation of the proposed attention module (plus visualizations).
\end{itemize}

\section{Related Work}

We first briefly review recent progresses in multi-label image classification.

Many methods focus on the semantic relationship between objects or object classes, by using dependency networks~\cite{2011_IJCAI_Markov}, pairwise co-occurrence adjacency matrices~\cite{2019_ICCV_GCNre,2019_CVPR_GCN,2011_ICCV_Matrix}, or conditional graphs~\cite{2016_CVPR_condition_graph}. However, inference in graphs can be very difficult due to the exponential label space, and is usually approximated by Gibbs sampling. Pairwise statistics for first-order adjacency matrix construction has attracted much attention recently~\cite{2019_CVPR_GCN,2019_ICCV_GCNre,2020_AAAI_CMA}, mainly attributed to the popularity of the Graph Convolutional Network (GCN)~\cite{GCN_original}. But, co-occurrence statistics in a small training set is not reliable,  and can easily cause overfitting. Besides, high order relationship is beyond what GCN can represent. Recurrent Neural Network (RNN) has also been applied in various studies~\cite{2016_CVPR_RNNCNN,2017_CVPR_RNN,2018_TMM_RNN} to explore high order label dependencies~\cite{2018_TMM_RNN}. But, the effectiveness of RNN in multi-label tasks is yet to be proved. Also, training RNN usually requires dedicated hyperparameter tuning, making it unsuitable in real applications.

Generating object proposals~\cite{2015_PAMI_HCP,2018_ACM_KD,Wider,2016_ICIP_RCP} is another approach. Object proposals are generated by methods like EdgeBoxes~\cite{EdgeBox} or Selective Search~\cite{SS}, then sent to a shared CNN. Finally, a category-wise max-pooling is used to fuse the proposals' scores. Proposals are, however, large in quantity and expensive in computing.

Attention mechanism is widely adopted in a variety of vision tasks, such as detection~\cite{2019_Thundernet} and tracking~\cite{2018_tracking_end,2018_deep_tracking}. In multi-label recognition, one representative paradigm is SRN~\cite{2017_CVPR_SRN}, which used a spatial regularization network to rectify original predictions. Sarafianos \etal~\cite{2018_ECCV_CAM} proposed a similar pipeline to aggregate visual attention, and Durand \etal~\cite{2017_VOC95} generated multiple class heat maps to assemble the prediction scores. These attention methods require deliberate design processes, and cannot be intuitively interpreted. Recently, You \etal~\cite{2020_AAAI_CMA} used a cross-domain model to extract class-specific features, but they dropped the class-agnostic average pooling. Gao and Zhou~\cite{2020_arxiv_Gaobb} proposed a detection-like pipeline for multi-label image classification, but the inference process is very expensive. There are other complicated attention-based methods, such as knowledge distillation~\cite{2018_ACM_KD}, visual attention consistency~\cite{2019_CVPR_VA}, and suppressing negative class activation map~\cite{2020_ECCV}.

Unlike existing attention-based methods that either drop the class-agnostic average pooling or design a complicated and nonintuitive pipeline, this paper proposes a residual attention module, which uses a class-agnostic average pooling and a class-specific spatial pooling to get robust features for multi-label image classification. We want to emphasize that our CSRA is significantly different from the previous attention model SRN~\cite{2017_CVPR_SRN}. CSRA reuses the classifier's weights and has no additional parameters, while SRN has a complicated pipeline and many extra parameters. We also generalize CSRA to multi-head attention, which is totally different from the whole structure in SRN. Our CSRA is end-to-end trainable, while SRN requires a three stage dedicated model training and finetuning process.

\section{Class-specific Residual Attention}

To present the proposed CSRA module, we start from the code in Fig.~\ref{fig:pytorch code} and the results in Table~\ref{tab:free improvement}.

\subsection{Why global max pooling helps?}

Table~\ref{tab:free improvement} lists experimental results of various backbone networks on four different datasets. Because only very few previous studies on multi-label recognition release their code or pretrained models, we also trained our own models to obtain baseline results using MobileNet~\cite{MobileNet}, ResNet~\cite{ResNet}, EfficientNet~\cite{Efficientnet} and VIT~\cite{VIT}, in addition to the GCN~\cite{2019_CVPR_GCN} method. In our simple modification, $\lambda$ was chosen in two ways. In the first (``Varying $\lambda$''), $\lambda$ was tuned for every experiment; while in the second (``Fixed $\lambda$''), we always used 0.02 for models in the ResNet family, and 0.2 for all other models. For multi-label tasks, we used mAP as the evaluation metric, and for ImageNet (which is single-label), accuracy (``acc'') was used.

In Table~\ref{tab:free improvement}, ``ResNet-cut'' was ResNet-101 pretrained on ImageNet with CutMix~\cite{cutmix}; pretrained ``ResNet-50/-101'' were downloaded from the PyTorch official website. ``EfficientNet'' and ``MobileNet'' were EfficientNet-B3~\cite{Efficientnet} and MobileNet-V2~\cite{MobileNet}, respectively. All the VIT~\cite{VIT} models were pretrained on ImageNet-21k and finetuned on ImageNet with the 224$\times$224 resolution, except ``VIT-B16-384'' (384$\times$384 resolution). For VIT~\cite{VIT} models, we dropped the class token and use the final output patch embeddings as the feature tensor, and we interpolated positional embeddings suggested in~\cite{VIT} when finetuning them with higher resolutions on VOC2007~\cite{VOC} and MS-COCO~\cite{MSCOCO}.

These results show that simply adding a max pooling can consistently improve multi-label recognition, especially when the baseline model's mAP is not high. In fact, as shown by the ``Fixed $\lambda$'' results, our simple modification is \emph{robust to $\lambda$}, which means that these 4 lines of code is \emph{already a practical method to improve multi-label recognition}: simple, minimal-cost, training-free, and insensitive to the hyperparameter.

Why the simple global max pooling helps? First, note that \pyth{y_max} in Fig.~\ref{fig:pytorch code} finds the maximum value among all spatial locations for each category. Hence, it can be viewed as a class-specific attention mechanism. We conjecture \emph{it focuses our attention to classification scores at different locations for different object categories}. This attention mechanism is, intuitively, very useful for multi-label recognition, especially when there are objects from many classes and with varying sizes.

\subsection{Residual attention}

The attention interpretation of Fig.~\ref{fig:pytorch code} inspires us to generalize it and design a trainable attention mechanism.

For a given image $I$, it is first sent to a feature extractor (the CNN backbone) $\phi$ to obtain a feature tensor $\mathbf{x}\in \mathbb{R}^{d\times h \times w}$, where $d$, $h$, $w$ are the dimensionality, height, width of the feature tensor:
\begin{equation}
	\mathbf{x} = \phi(I;\theta) \,,
\end{equation}
in which $\theta$ is the parameters of the CNN backbone. For simplicity, we adopt ResNet-101~\cite{ResNet} and input image resolution $224 \times 224$ as an example, if not otherwise specified. As a result, the shape of the feature tensor $\mathbf{x}$ is $2048\times7\times7$, which can be decoupled as $\mathbf{x}_1,\mathbf{x}_2,\dots,\mathbf{x}_{49}$ ($\mathbf{x}_i\in\mathbb{R}^{2048}$). Next, a fully connected ($1 \times 1$ convolution) layer is the classifier (\pyth{FC} in Fig.~\ref{fig:pytorch code}), with $\mathbf{m}_i\in\mathbb{R}^{2048}$ being the classifier for the $i$-th class. Constants such as 49 can be changed accordingly when a different setup has been used.

Now we define the \emph{class-specific attention scores} for the $i$-th class and $j$-th location as
\begin{equation} 
	\label{eq:sij}
	s_j^i = \frac{\exp(T \mathbf{x}_j^T \mathbf{m}_i )}{\sum_{k=1}^{49} \exp(T \mathbf{x}_k^T \mathbf{m}_i) } \,,
\end{equation}
where $\sum_{j=1}^{49} s_j^i=1$ and $T>0$ is the temperature controlling the scores' sharpness. We can view $s_j^i$ as the probability of the class $i$ appearing at location $j$.

Then, we can define the class-specific feature vector for class $i$ as a weighted combination of the feature tensor, where the attention scores for the $i$-th class $s_k^i$ $(1 \le k \le 49)$ are the weights, as
\begin{equation} 
	\label{eq:ai}
	\mathbf{a}^i = \sum_{k=1}^{49} s_k^i \mathbf{x}_k \,.
\end{equation}

In contrast, the classical global class-agnostic feature vector for the entire image is
\begin{equation} 
	\label{eq:g}
	\mathbf{g} = \frac{1}{49}\sum_{k=1}^{49} \mathbf{x}_k \,.
\end{equation}

\begin{figure}
	\centering
	\includegraphics[width=1.\columnwidth]{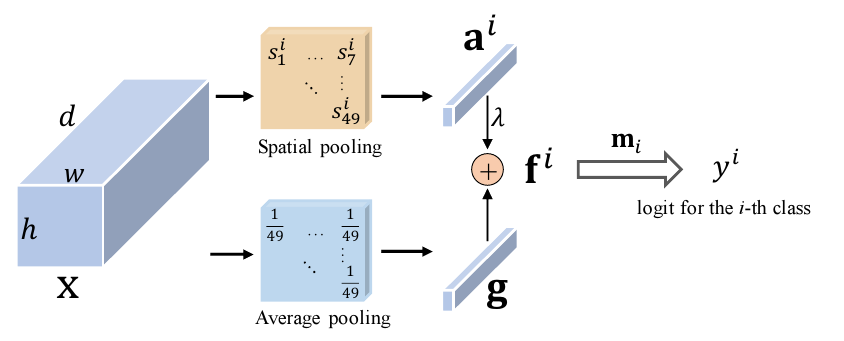}
	\caption{The proposed CSRA module to obtain features and classification results for the $i$-th class.}
	\label{fig:CSRA}
\end{figure}

Since $\mathbf{g}$ has been widely used and has achieved good results, we treat it as the main feature vector. And we treat $\mathbf{a}^i$ as class-specific \emph{residual} features. As illustrated in Fig.~\ref{fig:CSRA}, by adding these two vectors, we obtain our \emph{class-specific residual attention} (CSRA) feature $\mathbf{f}^i$ for the $i$-th class:
\begin{equation} 
	\label{eq:fi}
	\mathbf{f}^i = \mathbf{g} + \lambda \mathbf{a}^i \,.
\end{equation}
This constitutes the proposed CSRA module.

Finally, all these class-specific feature vectors are sent to the classifier to obtain the final logits
\begin{equation} 
	\hat{\mathbf{y}} \triangleeq (y^1, y^2, \dots, y^C) = (\mathbf{m}_1^T\mathbf{f}^1, \mathbf{m}_2^T\mathbf{f}^2, \dots, \mathbf{m}_C^T\mathbf{f}^C) \,,
	\label{eq:logits}
\end{equation}
where $C$ means the number of classes.

\subsection{Explaining the CSRA module}

We first prove that Fig.~\ref{fig:pytorch code} is in fact a special case of our CSRA module. By substituting Eq.~(\ref{eq:sij}) till Eq.~(\ref{eq:fi}) into Eq.~(\ref{eq:logits}), we can easily derive the logit for the $i$-th class as:
\begin{align} 
	y^i & = \mathbf{m}_i^T \mathbf{g} + \lambda \mathbf{m}_i^T \sum_{k=1}^{49} s_k^i \mathbf{x}_k 	\label{eq:yi} \\
		& = \frac{1}{49} \sum_{k=1}^{49} \mathbf{x}_k^T\mathbf{m}_i + \lambda \sum_{k=1}^{49}
		\frac{\exp (T \mathbf{x}_k^T \mathbf{m}_i)}{ \sum_{l=1}^{49} \exp(T\mathbf{x}_l^T\mathbf{m}_i)} \mathbf{x}_k^T \mathbf{m}_i \,. \label{eq:yi_softmax}
\end{align}
The first term in the right hand side of Eq.~(\ref{eq:yi}) is the base logit for the $i$-th class, which can be written as $\frac{1}{49}\sum_{k=1}^{49} \mathbf{x}_k^T \mathbf{m}_i$ (\pyth{y_avg} in Fig.~\ref{fig:pytorch code}). Inside the second term, $\mathbf{x}_k^T\mathbf{m}_i$ is the classification score at location $k$ ($1 \le k \le 49$) for the $i$-th class, which is then weighed by the normalized attention score $s_k^i$ to form the \emph{average} class-specific score.

It is well-known that when $T\rightarrow \infty$, the softmax output $s_k^i = \frac{\exp (T \mathbf{x}_k^T \mathbf{m}_i)}{ \sum_{l=1}^{49} \exp(T\mathbf{x}_l^T\mathbf{m}_i)}$ becomes a Dirac delta function, in which the maximum element in $s_k^i$ ($1 \le k \le 49$) corresponds to all the probability mass while all other elements are 0. Hence, when $T\rightarrow \infty$,
\begin{equation} 
	\label{eq:meanmax}
	y^i  = \mathbf{m}_i^T \mathbf{g} + \lambda \max (\mathbf{x}_1^T\mathbf{m}_i,\dots,\mathbf{x}_k^T\mathbf{m}_i) \,.
\end{equation}

Obviously, the last term in Eq.~(\ref{eq:meanmax}) exactly corresponds to \pyth{Lambda *}\pyth{ y_max} in Fig.~\ref{fig:pytorch code}, and the testing-time modification in Fig.~\ref{fig:pytorch code} is not only the motivation for, but also a special case of CSRA.

Comparing Eq.~(\ref{eq:meanmax}) with Eq.~(\ref{eq:yi_softmax}), instead of solely relying on one location for the residual attention, CSRA hinges on residual attention features from all locations. Intuitively, when there are multiple small objects from the same class in the input image, CSRA has a clear edge over global average pooling alone or global max pooling alone.

More specifically, we have
\begin{align} 
	\label{eq:fi_re}
	\mathbf{f}^i & = \mathbf{g} + \lambda \mathbf{a}^i                                                      \\
		         & =  \sum_{k=1}^{49} (\frac{1}{49} + \lambda s_k^i) \mathbf{x}_k                           \\
				 & = (1+\lambda)\sum_{k=1}^{49} \frac{\frac{1}{49} + \lambda s_k^i}{1+\lambda} \mathbf{x}_k
\end{align}
where $\sum_{k=1}^{49}\frac{1}{49}=1$ and $\sum_{k=1}^{49}s_k^i=1$, and the constant multiplier $1+\lambda$ can be safely ignored. Hence, the final CSRA feature vector for the $i$-th class is a weighted combination of the feature tensor $\mathbf{x}_k$. The weight for location $k$, $\frac{\frac{1}{49} + \lambda s_k^i}{1+\lambda}$, is normalized, because $\sum_{k=1}^{49} \frac{\frac{1}{49} + \lambda s_k^i}{1+\lambda} = 1$. This weight is also a weighted combination of two terms: $\frac{1}{49}$ and $s_k^i$:
\begin{itemize}
	\item $\frac{1}{49}$ corresponds to a prior term, which is independent of either category $i$ or location $k$;
	\item $s_k^i$, the probability of location $k$ occupied by an object from category $i$ (computed based on $\mathbf{x}_k$), corresponds to a data likelihood term, which hinges on both the classifier $\mathbf{m}_i$ for the $i$-th class and location $k$'s feature $\mathbf{x}_k$. From this perspective, it is both intuitive and reasonable to use this class-specific and data-dependent score as our attention score.
\end{itemize}

\subsection{Multi-head attention}\label{sec:MHA}

\begin{figure*}
	\centering
	\includegraphics[width=0.9\linewidth]{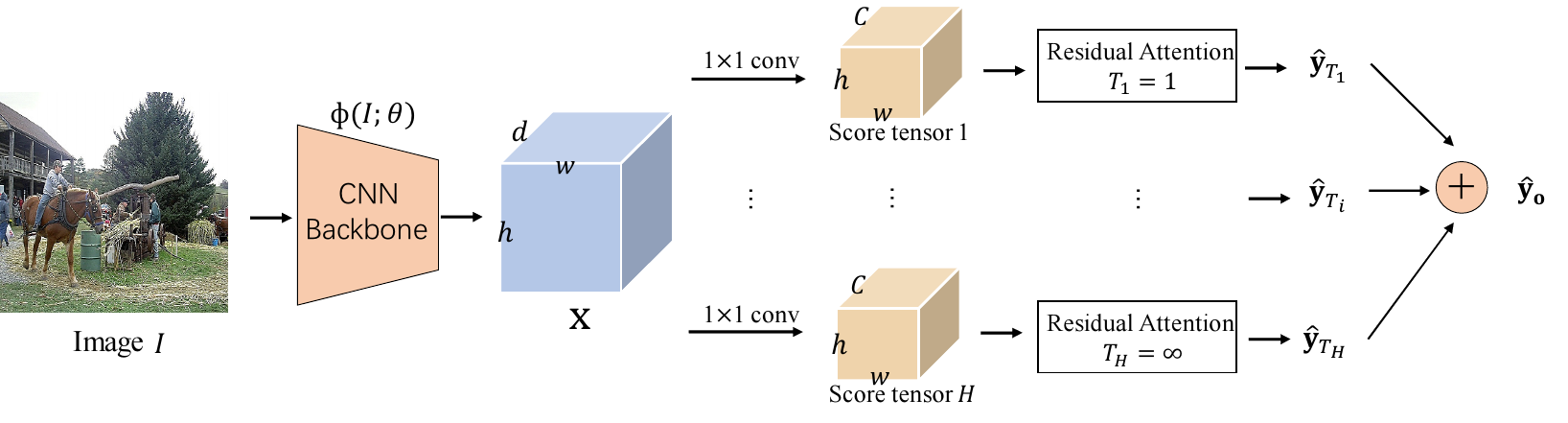}
	\caption{Overall pipeline of multi-head CSRA. An Image is first sent to a CNN backbone to get the feature tensor $\mathbf{x}$, which is used to generate multiple score tensors ($\in \mathbb{R}^{C\times h\times w}$, $C$ is the number of classes) by different $1 \times 1$ convolutions (FCs). The residual attention defined in Eq.~(\ref{eq:yi_softmax}) is applied to every score tensor to produce different logits $\hat{\mathbf{y}}_{T_i}$ ($i\in\{1,2,\dots,H\}$, $\hat{\mathbf{y}}_{T_i}\in \mathbb{R}^C$), which are then fused to get the final logits $\hat{\mathbf{y}}_o$. The temperature $T$ is different in different branches, but the same $\lambda$ is shared among them.}
	\label{fig:MHA}
\end{figure*}

The temperature hyperparameter $T$ may be tricky to tune, and it is possible that different classes may need different temperatures (or different attention scores). Thus, we further propose a simple multi-head attention extension to CSRA, as shown in Fig.~\ref{fig:MHA}.

Multiple branches (or heads) of residual attention (Fig.~\ref{fig:CSRA} / Eq.~\ref{eq:yi}) are used, with each branch utilizing a different temperature $T$ but sharing the same $\lambda$. We denote the number of attention heads as $H$. To rid ourselves of tuning the temperature $T$, we either choose single head ($H=1$) with a fixed temperature $T=1$, or use multi-head attention ($H>1$) with fixed sequences of temperatures $T_1, T_2, \dots, T_H$. Besides $H=1$, we also used $H=2,4,6,8$. Specifically, 
\begin{itemize}
	\item When $H=2$, $T_1=1$ and $T_2=\infty$ (i.e., max pooling); 
	\item When $H=4$, $T_{1:3}=1,2,4$ and $T_4 = \infty$; 
	\item When $H=6$, $T_{1:5}=1,2,3,4,5$ and $T_6=\infty$; 
	\item When $H=8$, $T_{1:7}=1,2,3,4,5,6,7$ and $T_8=\infty$. 
\end{itemize}
That is, when $H>1$, the final $T_H$ is always $\infty$, and the other $T$ are chosen in an increasing order. Different $T$ values can bring diversity into the branches, thus producing better classification results. In short, \emph{there is no need to tune $T$ in CSRA}.

For better convergence speed, we choose to normalize our classifier's weights to unit vectors (i.e., $\mathbf{m}_i \leftarrow \frac{\mathbf{m}_i}{||\mathbf{m}_i||}$). We will show empirically that this normalization makes no difference in accuracy, but it can lead to faster convergence in the training process.

The logits $\hat{\mathbf{y}}_{T_1},\hat{\mathbf{y}}_{T_2},\dots,\hat{\mathbf{y}}_{T_H}$ ($\hat{\mathbf{y}}_{T_i} \in \mathbb{R}^C$) from different heads are added to get the final logits, $\hat{\mathbf{y}}_{o}$ , as
\begin{equation}
	\hat{\mathbf{y}}_{o} = \sum_{h=1}^H \hat{\mathbf{y}}_{T_i} \,.
\end{equation}
where $T_i$ is the temperature for the $i$-th head.

Finally, the classical binary cross entropy (BCE) loss is used to compute the loss incurred between our prediction $\hat{\mathbf{y}}_{o}$ and the groundtruth labels, and the stochastic gradient descent (SGD) optimization method is used to minimize this loss function. Hence, the proposed CSRA (either single- or multi-head) is simple in structure and easy to implement.

\section{Experimental Results}

Now, we validate the effectiveness of CSRA and analyze its components empirically. We first describe the general experimental settings, then present our experimental results and compare CSRA with previous state-of-the-art models. Finally, we empirically analyze how the components and hyperparameters influence the performance of our model. We experimented with 4 multi-label datasets: VOC2007~\cite{VOC}, VOC2012~\cite{VOC2012}, MS-COCO~\cite{MSCOCO} and WIDER-Attribute~\cite{Wider}.

\subsection{Experimental settings}

\noindent\textbf{Training details} As aforementioned, we build multi-label recognition models in an end-to-end way by minimizing the binary cross entropy loss using SGD. For data augmentation, we only perform random horizontal flip and random resized crop, following previous work~\cite{2019_CVPR_GCN,2019_ICCV_GCNre}. When we train baseline models (BCE loss without CSRA), we use an initial learning rate of 0.01 for both the backbone and the classifiers. For training our residual attention models, we choose the learning rate of 0.1 for the CSRA module and classifiers, and 0.01 for the CNN backbone, respectively. We apply the warmup scheduler for training both the baseline models and our CSRA models. The CNN backbone is initialized from various pretrained models, and fine-tuned for 30 epochs on multi-label datasets. The momentum is 0.9, and weight decay is 0.0001. The batch size and input image resolution for the WIDER Attribute dataset ~\cite{Wider} are 64 and $224 \times 224$, respectively. For MS-COCO~\cite{MSCOCO}, VOC2007~\cite{VOC} and VOC2012~\cite{VOC2012}, the batch size and input image resolution are 16 and $448 \times 448$, respectively.

\vspace{6pt}\noindent\textbf{Evaluation metrics} The widely used mean average precision (mAP) is our primary evaluation metric. We set positive threshold as 0.5 and also adopt the overall precision (OP), overall recall (OR), overall F1-measure (OF1), per-category precision (CP), per-category recall (CR) and per-category F1-measure (CF1), following previous multi-label image classification research~\cite{2020_arxiv_Gaobb,2019_CVPR_GCN,2019_ICCV_GCNre,2019_CVPR_VA}.

\subsection{Comparison with state-of-the-arts}

\noindent\textbf{VOC2007} VOC2007~\cite{VOC} is a widely used multi-label image classification dataset. It has 9,963 images and 20 classes, in which the \emph{train-val} set has 5,011 images and the \emph{test} set has 4,952 images. We use the \emph{train-val} set for training and the \emph{test} set for evaluation. The input resolution is $448 \times 448$.

CSRA was applied to two backbones: the original ResNet-101, and ResNet-cut pretrained on ImageNet with CutMix~\cite{cutmix}. We also report the mAP of these backbones pretrained on the MS-COCO~\cite{MSCOCO} dataset. For simplicity, we only use one branch, i.e., $H=1$, $T=1$, and $\lambda=0.1$. As shown in Table~\ref{tab:voc2007}, CSRA surpasses previous state-of-the-art models.

\begin{table}
	\caption{Comparisons of mAP (in \%) of state-of-the-art models and our CSRA on VOC2007, where ``-'' means the results were not provided. ``ResNet-101'' is pretrained and downloaded from the PyTorch official website, ``ResNet-cut'' is ResNet-101 pretrained on ImageNet with CutMix~\cite{cutmix}. The baseline result of ResNet-101 is from~\cite{2020_arxiv_Gaobb}. ``extra data'' means pretrained on MS-COCO. The $^+$ symbol means using larger input image resolution.}
	\label{tab:voc2007}
	\centering
	\small
	\begin{tabular}{c|cc}
		\hline
		Method                       & mAP           & \thead{mAP\\(extra data)} \\
		\hline\hline
		RCP~\cite{2016_ICIP_RCP}     & 92.5          & -                              \\
		SSGRL$^+$~\cite{2019_ICCV_GCNre} & 93.4          & 95.0                           \\
		GCN~\cite{2019_CVPR_GCN}     & 94.0          & -                              \\
		ASL~\cite{ASL_2020_arxiv}    & 94.6          & 95.8                           \\
		\hline
		ResNet-101                   & 92.9          & -                              \\
		ResNet-cut                   & 93.9          & -                              \\
		ResNet-101 + CSRA              & \textbf{94.7} & \textbf{96.0}                  \\
		ResNet-cut + CSRA              & \textbf{95.2} & \textbf{96.8}                  \\
		\hline
	\end{tabular}
\end{table}

\vspace{6pt}\noindent\textbf{VOC2012} VOC2012~\cite{VOC2012} contains 11,540 \emph{train-val} images and 10,991 \emph{test} images. We train our model on the \emph{train-val} sets, and evaluate its performance on the official evaluation server. The settings are the same as those in VOC2007: $H=1$, $T=1$, $\lambda=0.1$ and $448 \times 448$ resolution. As shown in Table~\ref{tab:voc2012}, when only using ResNet-101 ImageNet pretrained models, our CSRA has surpassed previous methods. When pretrained on extra data (MS-COCO), the performance of CSRA can be further improved, achieving new state-of-the-art performance.

\begin{table}
	\caption{Comparisons of mAP (in \%) on the VOC2012 dataset. The $^+$ symbol means using larger input image resolution.}
	\label{tab:voc2012}
	\centering
	\small
	\begin{tabular}{c|c|c}
		\hline
		Method                                  & mAP           & \thead{mAP\\(extra data)} \\
		\hline\hline
		HCP~\cite{2015_PAMI_HCP}                & 90.5          & -                              \\
		RCP~\cite{2016_ICIP_RCP}                & 92.2          & -                              \\
		Fev+Lv~\cite{Fev_Lv}                    & 89.4          & -                              \\
		SSGRL$^+$~\cite{2019_ICCV_GCNre}        & 93.9          & 94.8                           \\
		\hline
		ResNet-101 + CSRA                       & \textbf{94.1} & \textbf{95.2}                  \\
		ResNet-cut + CSRA                       & \textbf{94.6} & \textbf{96.1}                  \\
		\hline
	\end{tabular}
\end{table}

\vspace{6pt}\noindent\textbf{MS-COCO} Microsoft COCO~\cite{MSCOCO} is widely used for segmentation, classification, detection and captioning. We use COCO-2014 in our experiments, which has 82,081 training and 40,137 validation images and 80 object classes. We train our model using three pretrained CNN backbones (ResNet-101, ResNet-cut and VIT-L16) on the \emph{train} set and evaluate them on the \emph{val} set. Following~\cite{2019_ICCV_GCNre,2019_CVPR_VA}, we report the precision, recall and F1-measure with and without Top-3 scores.

Note that the variation of objects' shapes and sizes are more complicated on MS-COCO than those in both VOC2007 and VOC2012, thus we adopt a larger trade-off parameter $\lambda$ with multiple heads for better attention. Specifically, when running ResNet-101 and ResNet-cut models, we adopt six attention heads ($H=6$), and choose $\lambda=0.5$ and $\lambda=0.4$, respectively. For the VIT-L16 backbone~\cite{VIT}, we adopt eight heads ($H=8$, $\lambda=1.0$).

The results are shown in Table~\ref{tab:mscoco}, in which the upper block lists results of methods using ResNet series models as backbones, and the lower block is for other backbones. It can be seen that when our CSRA module is added to the ResNet-101 model, there is a significant gain, raising mAP from 79.4\% to 83.5\%, a total of 4.1\% improvement. ResNet-cut (pretrained with CutMix) plus CSRA has achieved 85.6\% mAP, surpassing previous state-of-the-art models by a large margin. 

It is worth mentioning that previous methods such as MCAR~\cite{2020_arxiv_Gaobb} and KSSNet~\cite{2018_ACM_KD} used complicated and time consuming pipelines. In comparison, our residual attention model is not only effective, but also surprisingly simple.

When running non-ResNet series models, we choose VIT-L16 pretrained on ImageNet with 224$\times$224 input and finetune it on MS-COCO with 448$\times$448 resolution (we interpolate the positional embeddings as suggested in~\cite{VIT}). In comparison with ASL~\cite{ASL_2020_arxiv} that used TResNet~\cite{Tresnet}, our residual attention model, VIT-L16 + CSRA, has achieved state-of-the-art performance by increasing mAP from 80.4\% to 86.5\%, a significant improvement of 6.1\%. Note that ASL~\cite{ASL_2020_arxiv} used multiple complex data augmentation methods, such as Cutout~\cite{cutout}, GPU Augmentations~\cite{ASL_2020_arxiv} or RandAugment~\cite{randaugment}, while we only used classic simple data augmentation (horizontal flip and random resize crop). When we use the RandAugment~\cite{randaugment} as our data augmentation technique, our CSRA was further improved to 86.9\% mAP (denoted as VIT-L16 + CSRA$^*$ in Table~\ref{tab:mscoco}) on MS-COCO.

\begin{table*}
	\caption{Comparisons of mAP (in \%) and multiple other metrics on MS-COCO. The upper block corresponds to ResNet-101-based models, and the lower block is for other non-ResNet models. The $^*$ symbol means using RandAugment~\cite{randaugment}, which was used in ASL~\cite{ASL_2020_arxiv}. The highest scores in each block are shown in boldface.}
	\label{tab:mscoco}
	\centering
	\small
	\begin{tabular}{c|c|clllll | lllllc}
		\hline
		&      & \multicolumn{6}{c|}{All} & \multicolumn{6}{c}{Top 3}                  \\ 
		\hline
		Methods                      & mAP           & CP                       & CR                        & CF1           & OP            & OR            & OF1           & CP            & CR            & CF1           & OP            & OR            & OF1           \\ \hline\hline
		ResNet-101                    & 79.4          & 83.4                     & 66.6                      & 74.0          & 86.8          & 71.1          & 78.2          & 86.2          & 59.7          & 70.6          & 90.5          & 63.7          & 74.8          \\
		ResNet-cut                    & 82.1          & 86.2                     & 68.7                      & 76.4          & \textbf{88.9}  & 73.1          & 80.3          & 88.7          & 61.3          & 72.5          & 92.1          & 65.2          & 76.3          \\
		ML-GCN~\cite{2019_CVPR_GCN}  & 83.0          & 85.1                     & 72.0                      & 78.0          & 85.8          & 75.4          & 80.3          & 89.2          & 64.1          & 74.6          & 90.5          & 66.5          & 76.7          \\
		MS-CMA~\cite{2020_AAAI_CMA}  & 83.8          & 82.9                     & 74.4                      & 78.4          & 84.4          & 77.9          & 81.0          & 88.2          & 65.0          & 74.9          & 90.2          & 67.4          & 77.1          \\
		KSSNet~\cite{2018_ACM_KD}    & 83.7          & 84.6                     & 73.2                      & 77.2          & 87.8          & 76.2          & 81.5          & -             & -             & -             & -             & -             & -             \\
		MCAR~\cite{2020_arxiv_Gaobb} & 83.8          & 85.0                     & 72.1                      & 78.0          & 88.0          & 73.9          & 80.3          & 88.1          & 65.5          & 75.1          & 91.0          & 66.3          & 76.7          \\
		ResNet-101 + CSRA          & 83.5          & 84.1                     & 72.5                      & 77.9          & 85.6          & 75.7          & 80.3          & 88.5          & 64.2          & 74.4          & 90.4          & 66.4          & 76.5          \\
		ResNet-cut + CSRA          & \textbf{85.6} & \textbf{86.2}            & \textbf{74.9}             & \textbf{80.1} & 86.6          & \textbf{78.0} & \textbf{82.1} & \textbf{90.1} & \textbf{65.7} & \textbf{76.0} & \textbf{91.4} & \textbf{67.9} & \textbf{77.9} \\
		\hline\hline
		ASL~\cite{ASL_2020_arxiv}    & 86.5          & 87.2                     & \textbf{76.4}            & \textbf{81.4}     & 88.2       & \textbf{79.2}          & 81.8          & 91.8          & 63.4          & 75.1          & 92.9       & 66.4          & 77.4          \\
		VIT-L16                      & 80.4          & 83.8                     & 67.0                      & 74.5          & 86.6          & 72.0          & 78.6          & 86.8          & 60.0          & 70.1          & 90.3          & 64.7          & 75.4          \\
		VIT-L16 + CSRA           & 86.5 & 88.2            & 74.4                      & 80.8          & 88.5 & 77.4          & 82.6 & 91.9 & \textbf{65.8} & 76.7 & 92.6          & \textbf{68.2} & 78.5 \\
		VIT-L16 + CSRA$^*$           & \textbf{86.9} & \textbf{89.1}            & 74.2                      & 81.0          & \textbf{89.6} & 77.1          & \textbf{82.9} & \textbf{92.5} & \textbf{65.8} & \textbf{76.9} & \textbf{93.4}          & 68.1 & \textbf{78.8} \\\hline
	\end{tabular}
\end{table*}

When we compare more specific metrics (such as CP, CR and CF1), the proposed CSRA method also has clear advantages.

\vspace{6pt}\noindent\textbf{WIDER-Attribute} WIDER-Attribute~\cite{Wider} is a pedestrian dataset containing 14 categories (human attributes) of each person. The training and validation sets have 28,345 annotated people and the test set has 29,179 people. Following the conventional settings, we use the \emph{train-val} set for training and evaluate the performance on the \emph{test} set. As this dataset has unspecified labels, we set them as negative in the training stage and ignore these unspecified labels in the test stage, following previous settings in~\cite{2017_CVPR_SRN}. We adopt VIT~\cite{VIT} as our backbone and evaluate its performance on this pedestrian dataset with or without the proposed CSRA module. For simplicity, we only use one attention head ($H=1$), and choose $\lambda=0.3$, $T=1$ in our CSRA model. The input image resolution is $224\times224$.

For running VIT-B16 and VIT-L16, we drop the class token and use the final patch embeddings as the feature tensor. As shown in Table~\ref{tab:wider}, the improvement of the backbone is large, from 86.3\% to 89.0\% in the VIT-B16 backbone and from 87.7\% to 90.1\% in the VIT-L16 backbone. Together with the experimental results on MS-COCO, we have shown that the proposed CSRA module is not only suitable for the classic ResNet backbones, but also for emerging non-convolutional deep networks like vision transformers.

\begin{table}
	\caption{Comparisons of mAP (in \%) of state-of-the-art models and our CSRA on the WIDER-Attribute dataset.}
	\label{tab:wider}
	\centering
	\small
		\begin{tabular}{c|c|c|l}
			\hline
			Methods                    & mAP           & CF1           & OF1           \\ \hline\hline
			DHC~\cite{Wider} & 81.3          & -             & -             \\
			VA~\cite{VA_wider}         & 82.9          & -             & -             \\
			SRN~\cite{2017_CVPR_SRN}   & 86.2          & 75.9          & 81.3          \\
			VAA~\cite{2018_ECCV_CAM}       & 86.4          & -             & -             \\
			VAC~\cite{2019_CVPR_VA}    & 87.5          & 77.6          & 82.4          \\
			\hline
			VIT-B16                    & 86.3          & 75.9          & 81.5          \\
			VIT-L16                    & 87.7          & 78.1          & 82.8          \\
			VIT-B16 + CSRA          & \textbf{89.0} & \textbf{79.4} & \textbf{84.3} \\
			VIT-L16 + CSRA          & \textbf{90.1} & \textbf{81.0} & \textbf{85.2} \\
			\hline
		\end{tabular}
\end{table}

\subsection{Effects of various components in CSRA}

Finally, we study the effects of various components in the proposed CSRA module.

\vspace{6pt}\noindent\textbf{Class-agnostic vs. class-specific} To further verify whether the class-agnostic average pooling matters much to the final performance, we did a controlled experiment, modifying only the process of calculating the overall feature for the $i$-th class. When we only apply the average pooling, the overall feature $\mathbf{f}^i=\mathbf{g}$, which is the same as the baseline method. When we apply only the spatial pooling, the overall feature $\mathbf{f}^i=\mathbf{a}^i=\sum_{k=1}^{49} s_k^i\mathbf{x}_k$. When the two are combined, $\mathbf{f}^i=\mathbf{g}+\lambda \mathbf{a}^i$, which is the proposed CSRA.

Table~\ref{tab:priori} shows the results of ResNet-cut with one attention head ($\lambda=0.4$) on the MS-COCO dataset. The class-specific attention (spatial pooling) is more effective than the class-agnostic average pooling. By combining the two, CSRA clearly outperforms both.

\begin{table}
	\caption{The effect of applying average pooling versus spatial pooling on the MS-COCO dataset.}
	\label{tab:priori}
	\centering
	\small
	\begin{tabular}{c|c|c|c|l}
		\hline
		Backbone                                        & Method & average & spatial & mAP           \\ \hline\hline
		\multirow{3}{*}{\thead{ResNet-cut\\$H=1$, $T=1$}} & 1      & $\surd$ &         & 82.1          \\ \cline{2-5}
		                                                & 2      &         & $\surd$ & 84.2          \\ \cline{2-5}
		                                                & 3      & $\surd$ & $\surd$ & \textbf{85.3} \\ \hline
	\end{tabular}
\end{table}

\begin{figure}
	\centering
	\subfigure[Original image] { \includegraphics[width=0.47\linewidth]{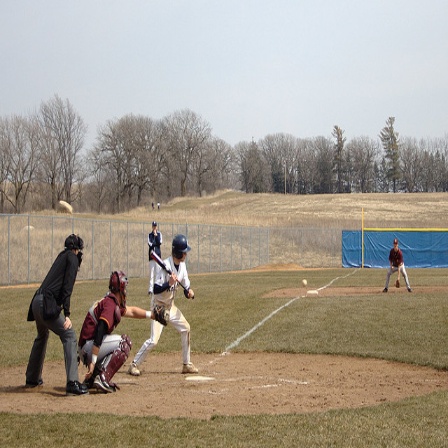} }
	\subfigure[Attention image] { \includegraphics[width=0.47\linewidth]{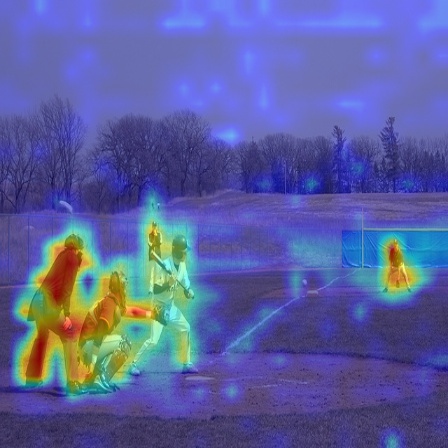} }
	\caption{One sample image from MS-COCO 2014 (on the left) and the attention scores for the `person' class overlapped on top of it (on the right). The score map was resized to the same size as that of the input image. }
	\label{fig:vis}
\end{figure}

\vspace{6pt}\noindent\textbf{Visualizing the attention} Table~\ref{tab:priori} confirms that class-specific attention is important for multi-label recognition. Intuitively, we believe that the proposed CSRA is in particular valuable when there are many small objects. We resort to visualization of the attention scores to verify this intuition, which are shown in Fig.~\ref{fig:vis}.

The visualization shows that even when there are multiple persons (including a person occupying only a small number of pixels), the attention score of CSRA ($s_j^i$ in Eq.~\ref{eq:sij}, with $i$ corresponds to the `person' class, and $j$ enumerates all locations) effectively captures where are the persons. 

In general, the score maps often accurately localize objects from different categories. More visualizations are shown in the supplementary material of this paper.

\begin{figure}
	\centering
	\subfigure[VIT-L16 ($H=8$) on MS-COCO] { \includegraphics[width=0.775\linewidth]{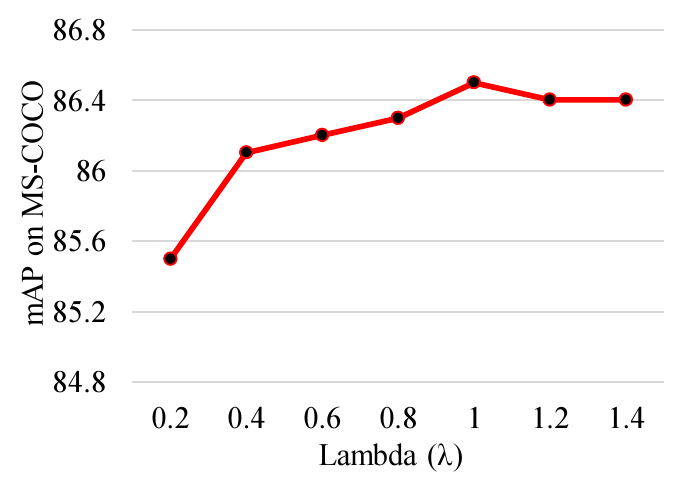} }
	\subfigure[ResNet-cut ($H=1$) on VOC2007]    { \includegraphics[width=0.775\linewidth]{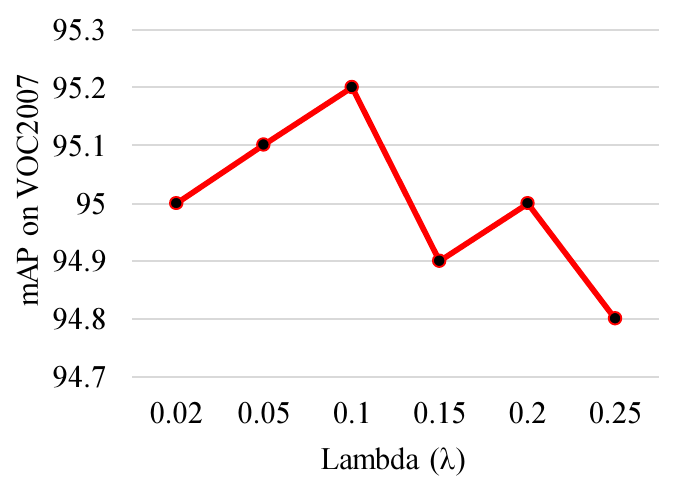} }
	\caption{The influence of $\lambda$ on the MS-COCO dataset using VIL-L16 as the backbone, and on VOC2007 using ResNet-cut. We set classifier num $H=8$ and $H=1$ for them, respectively.}
	\label{fig:lam}
\end{figure}

\vspace{6pt}\noindent\textbf{Effect of $\lambda$} Since we have fixed the sequence of temperature values, the only hyperparameter in CSRA is $\lambda$. We take VIT-L16 and ResNet-cut as our backbone networks, and evaluate the performance of different $\lambda$ on the MS-COCO and VOC2007 datasets. As show in Fig.~\ref{fig:lam}, the performance of VIL-L16 steadily increases to its peak around $\lambda=1.0$, while the ResNet-cut reaches its highest score at $\lambda = 0.1$. Fig.~\ref{fig:lam} suggests that CSRA is relatively robust to $\lambda$, because all these $\lambda$ values produce mAPs far higher than the baseline method. However, different backbone may need different $\lambda$.

When $\lambda$ becomes too large, the effect of the average pooling component becomes declining, and the spatial pooling will dominate our model. As shown in Table~\ref{tab:priori}, spatial pooling alone is inferior to CSRA (which combines average and spatial pooling). Hence, it is possible that a too large $\lambda$ will lead to a performance drop in our CSRA.

\vspace{6pt}\noindent\textbf{Number of attention heads} We then test how the number of attention heads is affecting the model's performance. Similarly, we use VIT-L16 and ResNet-cut as our backbone, and evaluate the performance of different number of heads, $H$. As shown in Table~\ref{tab:head}, the mAP increases steadily when $H$ increases to a large number (6 or 8), demonstrating the effectiveness of the multi-head attention version of CSRA.

\begin{table}
	\caption{The influence of the number of attention heads in CSRA tested on the MS-COCO dataset.}
	\label{tab:head}
	\centering
	\small
	\setlength{\tabcolsep}{3.2pt}
	\begin{tabular}{c|ccccc}
		\hline
		           & $H=1$ & $H=2$ & $H=4$ & $H=6$         & $H=8$         \\ \hline \hline
		VIT-L16    & 85.8  & 86.1  & 86.4  & 86.4          & \textbf{86.5} \\ \hline
		ResNet-cut & 85.3  & 85.4  & 85.5  & \textbf{85.6} & 85.5          \\ \hline
	\end{tabular}
\end{table}

It is also worth noting that even $H=8$ leads to very small computational overhead, thanks to the simplicity of our CSRA attention module. On MS-COCO, in the training stage the baseline method (without CSRA) took 3705.6 seconds, while CSRA with 8 heads took 3735.7 seconds, with only 0.8\% overhead. The total test time on MS-COCO increased from 142.3 to 153.8 seconds, and this increase (8\%) is acceptable when considering the significant improvement in mAP.

\vspace{6pt}\noindent\textbf{Normalization} To test the effect of the normalization of the classifiers $\mathbf{m}_i$ (Sec.~\ref{sec:MHA}), we adopt ResNet-101 as our backbone, and evaluate the performance on MS-COCO, because this dataset is assumed to be the most representative multi-label image classification dataset.

\begin{table}
	\caption{The influence of normalization of classifier on the MS-COCO dataset with the ResNet-101 CNN backbone. The numbers are mAPs under different settings.}
	\label{tab:normalize}
	\centering
	\small
	\begin{tabular}{c|c|c|l}
		\hline
		                            & Normalize & $H=2$         & $H=4$           \\ \hline\hline
		\multirow{2}{*}{ResNet-101} & with     & 83.2          & 83.3          \\ \cline{2-4}
		                            & without   & 83.1          & 83.4 \\ \hline
	\end{tabular}
\end{table}

Table~\ref{tab:normalize} shows the influence of normalization. Since all mAP results are close to each other, the effect of normalization is limited in terms of mAP (the difference is 0.1 in both $H=2$ and $H=4$). We use this normalization step not for higher accuracy, but to increase the convergence speed during training.

\section{Conclusions and Future Work}

In this paper, we proposed CSRA, a simple but effective pipeline for multi-label image classification. The inspiration for CSRA came from our simple modification in the testing stage, where 4 lines of code brought consistent improvement to a variety of existing models without training. We generalized this modification to capture separate features for every object class, leading to the proposed class-specific residual attention module. A multi-head attention version of CSRA not only improved recognition accuracy, but also removed the dependency on a hyperparameter. CSRA outperformed existing methods on 4 benchmark datasets, albeit being both simpler and more explainable.

In the future, we will generalize our method to more general image representation learning, and further verify whether our residual attention will be useful in other computer vision tasks, such as object detection.

{\small
\bibliographystyle{ieee_fullname}
\bibliography{egbib}
}

\end{document}